\documentclass[11pt,a4paper]{article}
\usepackage[hyperref]{naaclhlt2018}
\usepackage{times}
\usepackage{latexsym}

\usepackage{url}
\usepackage{graphicx}
\usepackage{amsmath}
\usepackage{enumerate}
\usepackage{amsfonts}
\usepackage{array}
\usepackage{subfigure}
\usepackage{amssymb}
\usepackage{multirow}
\usepackage{booktabs}
\usepackage{footmisc}
\usepackage{floatrow}
\usepackage{mdwlist}
\usepackage{paralist}
\usepackage[linesnumbered,ruled,vlined]{algorithm2e}

\aclfinalcopy 
\def\aclpaperid{408} 

\title{Reinforced Co-Training}

\author{Jiawei Wu \\
  Department of Computer Science \\
  University of California\\ 
  Santa Barbara, CA 93106 USA \\
  \href{mailto:jiawei_wu@cs.ucsb.edu}{\texttt{jiawei\_wu@cs.ucsb.edu}} \\\And
  Lei Li \\
  Toutiao AI Lab \\
  Bytedance Co. Ltd \\
  Beijing, 100080 China\\
  \href{mailto:lileicc@gmail.com}{\texttt{lileicc@gmail.com}} \\\And
  William Yang Wang\\
  Department of Computer Science\\
  University of California\\
  Santa Barbara, CA 93106 USA \\
  \href{mailto:william@cs.ucsb.edu}{\texttt{william@cs.ucsb.edu}} \\}

\date{}

\begin{document}
\maketitle

\begin{abstract}
Co-training is a popular semi-supervised learning framework to utilize a large amount of unlabeled data in addition to a small labeled set. Co-training methods exploit predicted labels on the unlabeled data and select samples based on prediction confidence to augment the training. 
However, the selection of samples in existing co-training methods is based on a predetermined policy, which ignores the sampling bias between the unlabeled and the labeled subsets, and fails to explore the data space.
In this paper, we propose a novel method, Reinforced Co-Training, to select high-quality unlabeled samples to better co-train on. More specifically, our approach uses Q-learning to learn a data selection policy with a small labeled dataset, and then exploits this policy to train the co-training classifiers automatically. Experimental results on clickbait detection and generic text classification tasks demonstrate that our proposed method can obtain more accurate text classification results.

\end{abstract}

\section{Introduction}
\label{sec:intro}
Large labeled datasets are often required to obtain satisfactory performance for natural language processing tasks. However, it is time-consuming to label text corpus manually. In the meanwhile, there are abundant unlabeled text corpora available on the web. Semi-supervised methods permit learning improved supervised models by jointly train on a small labeled dataset and a large unlabeled dataset~\cite{zhu2006semi,chapelle2009semi}.

\begin{figure}[!htbp]
\centering
\includegraphics[width=0.9\columnwidth]{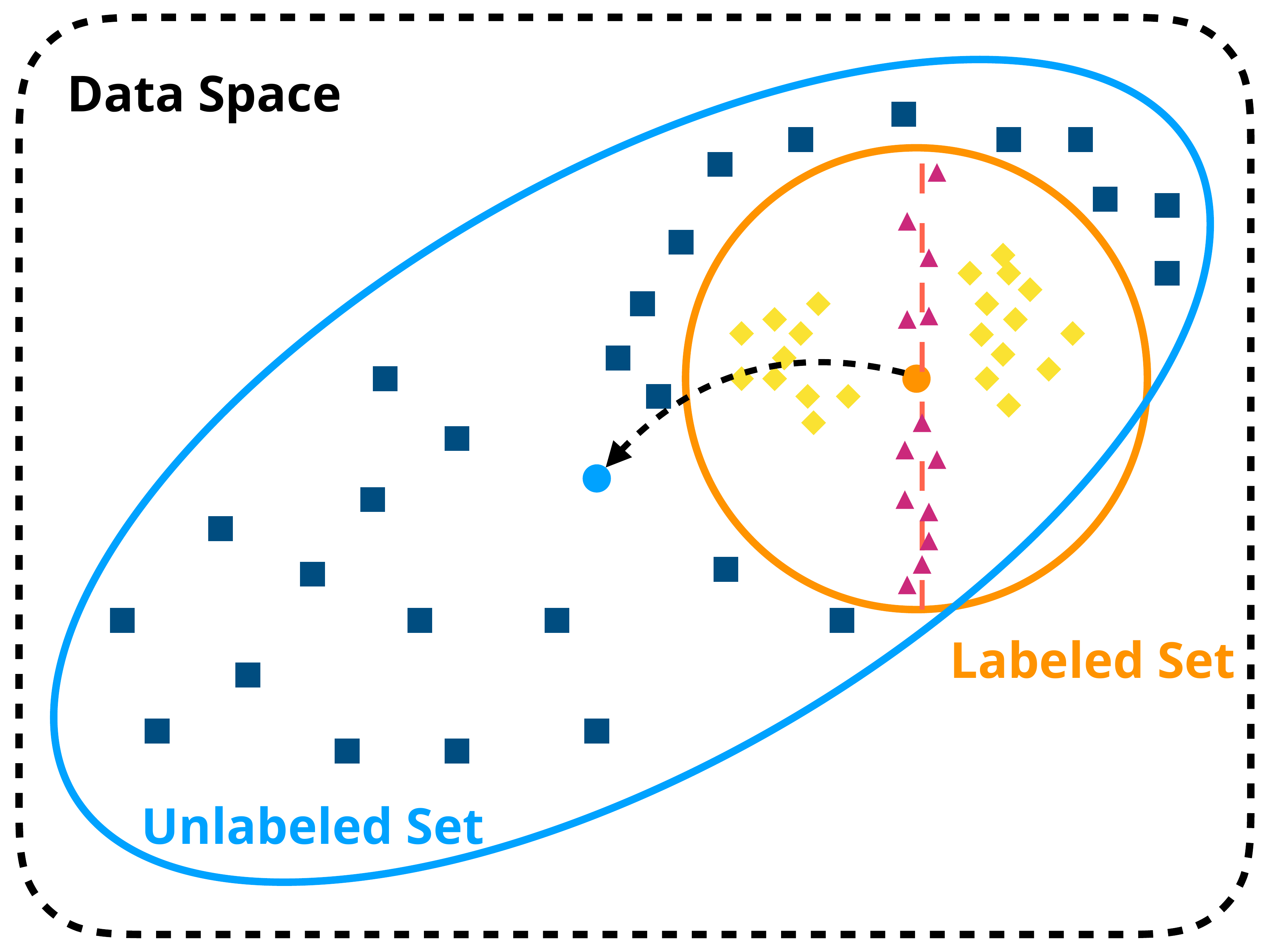}
\caption{Illustration of sample-selection issues in co-training methods. (1) Randomly sampled unlabeled examples ($\Box$) will result in high sampling bias, which will cause bias shift towards the unlabeled dataset ($\leftarrow$). (2) High-confidence examples ($\Diamond$) will contribute little during the model training, especially for discriminating the boundary examples ($\triangle$), resulting in myopic trained models.}\label{bias}
\end{figure}

Co-training is one of the widely used semi-supervised methods, where two complementary classifiers utilize large amounts of unlabeled examples to bootstrap the performance of each other iteratively \cite{blum1998combining,nigam2000analyzing}.
Co-training can be readily applied to NLP tasks since data in these tasks naturally have two or more views, such as multi-lingual data \cite{wan:2009:ACLIJCNLP} and document data (headline and content) \cite{ghani2000using,denis2003text}.
In the co-training framework, each classifier is trained on one of the two views (aka a subset of features) of both labeled and unlabeled data, under the assumption that either view is sufficient to classify. 
In each iteration, the co-training algorithm selects high confidence samples scored by each of the classifiers to form an auto-labeled dataset, and the other classifier is then updated with both labeled data and additional auto-labeled set. 
However, as shown in Figure \ref{bias}, most of existing co-training methods have some disadvantages. Firstly, the sample selection step ignores distributional bias between the labeled and unlabeled sets. 
It is common in practice to use unlabeled datasets collected differently from the labeled set, resulting in a significant difference in their sample distribution. 
After iterative co-training, the sampling bias may shift towards the unlabeled set, which results in poor performance of the trained model at the testing time. To remedy such bias, an ideal algorithm should select those samples according to the target (potentially unknown) testing distribution.
Secondly, the existing sample selection and training can be myopic. Conventional co-training methods select unlabeled examples with high confidence predicted by trained models. 
This strategy often causes only those unlabeled examples that match well to the current model being picked during iteration and the model might fail to generalize to complete sample space~\cite{zhang2006new}.
It relates to the well-known exploration-exploitation trade-off in machine learning tasks. An ideal co-training algorithm should explore the space thoroughly to achieve globally better performance. 
These intuitions inspire our work on learning a data selection policy for the unlabeled dataset in co-training.

The iterate data selection steps in co-training can be viewed as a sequential decision-making problem. To resolve both issues discussed above, we propose \textbf{Reinforced Co-Training}, a reinforcement learning (RL)-based framework for co-training.
Concretely, we introduce a joint formulation of a Q-learning agent and two co-training classifiers. In contrast to previous predetermined data sampling methods of co-training, we design a Q-agent to automatically learn a data selection policy to select high-quality unlabeled examples.
To better guide the policy learning of the Q-agent, we design a state representation to delivery the status of classifiers and utilize the validation set to compute the performance-driven rewards. Empirically, we indicate that our method outperforms previous related methods on clickbait detection and generic text classification problems. In summary, our main contributions are three-fold:
\begin{itemize}
\item We are first to propose a joint formulation of RL and co-training methods;
\item Our learning algorithm can learn a good data selection policy to select high-quality unlabeled examples for better co-training;
\item We show that our method can apply to large-scale document data and outperform baselines in semi-supervised text classification.
\end{itemize} 

In Section \ref{sec:related}, we outline related work in semi-supervised learning and co-training. We then describe our proposed method in Section \ref{sec:method}. We show experimental results in Section \ref{sec:experiments}. Finally, we conclude in Section \ref{sec:conclusion}.

\begin{figure*}[htbp]
\centering
\includegraphics[width=0.80\textwidth]{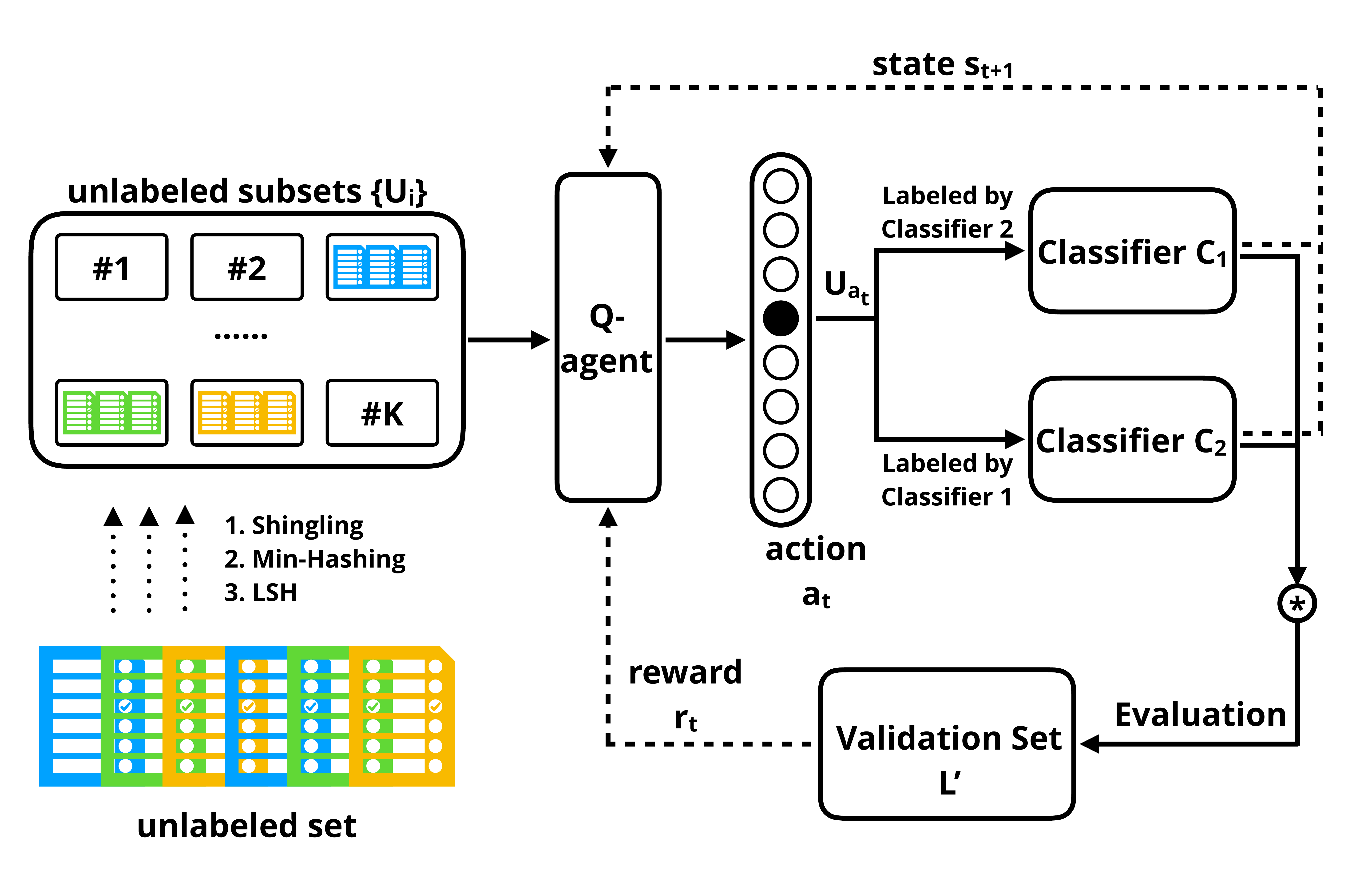}
\caption{The Reinforced Co-Training framework. \label{overview} }
\end{figure*}

\section{Related Work}
\label{sec:related}
Semi-supervised learning algorithms have been widely used in NLP \cite{liang2005semi}. As for text classification, \newcite{dai2015semi} introduce a sequence autoencoder to pre-train the parameters for the later supervised learning process. \newcite{johnson2015semi,johnson2016supervised} propose a method to learn embeddings of small text regions from unlabeled data for integration into a supervised convolutional neural network (CNN) or long short-term memory network (LSTM). \newcite{miyato2016adversarial} further apply perturbations to the word embeddings and pre-train the supervised models through adversarial training. However, these methods mainly focus on learning the local word-level information and pre-trained parameters from unlabeled data, which fails to capture the overall text-level information and potential label information.

Co-training can capture the text-level information of unlabeled data and generate pseudo labels during the training, which is especially useful on unlabeled data with two distinct views \cite{blum1998combining}. However, the confidence-based data selection strategies \cite{goldman2000enhancing,zhou2005tri,zhang2011cotrade} often focus on some special regions of the input space and fail to generate an accurate estimation of data space. \newcite{zhang2006new} proposes a performance-driven data selection strategy based on pseudo-accuracy and energy regularization. Meanwhile, \newcite{chawla2005learning} argues that the random data sampling method often causes sampling bias shift of the trained model towards the unlabeled set.

Comparing to previous related methods, our Reinforced Co-Training model can learn a performance-driven data selection policy to select high-quality unlabeled data. Furthermore, the performance estimation is more accurate due to the validation dataset and the data selection strategy is automatically learned instead of human designed. Lastly, the selected high-quality unlabeled data can not only help explore the data space but also reduce the sampling bias shift.

Our work is also related to recent studies in ``learning to learn''~\cite{maclaurin2015gradient,zoph2016neural,chen2017learning,wichrowska2017learned,yeung2017learning}. Learning to learn is one of the meta-learning methods~\cite{schmidhuber1987evolutionary,bengio1991learning}, where one model is trained to learn how to optimize the parameters of another certain algorithm. While previous studies focus more on neural network optimization~\cite{chen2017learning,wichrowska2017learned} and few-shot learning~\cite{vinyals2016matching,ravi2016optimization,finn2017model}, we are first to explore how to learn a high-quality data selection policy in semi-supervised methods, in our case, the co-training algorithm.

\section{Method}
\label{sec:method}
In this section, we describe our RL-based framework for co-training in detail. The conventional co-training methods follow the framework:
\begin{enumerate}
\item Initialize two classifiers by training on the labeled set;
\item Iteratively select a subset of unlabeled data based on a predetermined policy;
\item Iteratively update two classifiers with the selected subset of unlabeled data in addition to the labeled one.
\end{enumerate}
Step 2 is the core of different co-training variants. The original co-training algorithm is equipped with a policy of selecting high-confidence samples by two classifiers. Our main idea is to improve the policy by reinforcement learning. 

We formulate the data selection process as a sequential decision-making problem and the decision (action) $a_t$ at each iteration (time step) $t$ is to select a portion of unlabeled examples. This problem can be solved with an RL-agent by learning a policy. We first describe how we organize the large unlabeled dataset to improve the computational efficiency. Then we briefly introduce the classifier models used in co-training. After that, we describe the Q-agent, the RL-agent used in our framework and the environment in RL. The two co-training classifiers are integrated into the environment and the Q-agent can learn a good data selection policy by interacting with the environment. Finally, we describe how to train the Q-agent in our unified framework.

\subsection{Partition Unlabeled Data}
\label{subsec:partition}
Considering that the number of unlabeled samples is enormous, it is not efficient for the RL-agent to select only one example at each time step $t$. Thus, first we want to partition documents from the unlabeled dataset into different subsets based on their similarity. At each time step $t$, the RL-agent applies a policy to select one subset instead of one sample and then update the two co-training classifiers, which can significantly improve the computational efficiency.

Suppose each example in the unlabeled dataset as document $D$, where $D$ is the concatenation of the headline and paragraph. $V$ is the vocabulary of these documents. These documents are partitioned into different subsets based on Jaccard similarity, which is defined as:
\begin{equation}
sim(D_1,D_2) = \frac{|D_1\cap D_2|}{|D_1\cup D_2|},
\end{equation}
where $D_1, D_2 \in \mathbb{R}^{|V|}$ are the one-hot vectors of each document example.

Based on Jaccard similarity, the unlabeled examples can be split into different subsets using the following three steps, which have been widely used in large-scale web search \cite{rajaraman2010finding}: 
\begin{inparaenum}[1)]
\item Shingling, 
\item Min-Hashing, and
\item Locality-Sensitive Hashing (LSH).
\end{inparaenum}

After partition, the unlabeled set $U$ can be converted into $K$ different subset $\{U_1, U_2, ..., U_K\}$. Meanwhile, for each subset $U_i$, the first added document example $S_i$ is recorded as the representative example of the subset $U_i$. Choosing representative samples will help evaluate the classifiers on different subsets and obtain the state representations, which will be discussed in \ref{subsubsec:state}.

\subsection{Classifier Models}
\label{subsec:classifier}
As mentioned before, much linguistic data naturally has two or more views, such as multi-lingual data \cite{wan:2009:ACLIJCNLP} and document data (headline + paragraph) \cite{ghani2000using,denis2003text}. Based on the two views of data, we can construct two classifiers respectively. At the beginning of a training episode, the two classifiers are first seeded with a small set of labeled (seeding) training data $L$. At each time step $t$, the RL-agent makes a selection action $a_t$, and then the unlabeled subset $U_{a_t}$ is selected to train the two co-training classifiers. Following the standard co-training process \cite{blum1998combining}, at each time step $t$, the classifier $C_1$ annotate the unlabeled subset $U_{a_t}$ and the pseudo-labeled $U_{a_t}$ and the small labeled set $L$ are then used to update the classifier $C_2$, vice versa. In this way, we can boost the performance of $C_1$ and $C_2$ simultaneously.

\subsection{Q-Learning Agent}
\label{subsec:agent}
Q-learning is a widely used method to find an optimal action-selection policy \cite{watkins1992q}. The core of our model is a Q-learning agent, which is trained to learn a good policy to select high-quality unlabeled subsets for co-training. At each time step $t$, the agent observes the current state $s_t$, and selects an action $a_t$ from a discrete set of actions $A = \{1, 2,..., K\}$. Based on the action $a_t$, the two co-training classifiers $C_1$ and $C_2$ then can be updated with the unlabeled subset $U_{a_t}$ as described in Section \ref{subsec:classifier}. After that, the agent receives a performance-driven reward $r_t$ and the next state observation $s_{t+1}$. The goal of our Q-agent at each time step $t$ is to choose the action that can maximize the future discount reward
\begin{equation}
R_t = \sum_{t'=t}^{T}\gamma^{t'-t}r_{t'},
\end{equation}
where a training episode terminates at time $T$ and $\gamma$ is the discount factor.

\subsubsection{State Representation}
\label{subsubsec:state}
The state representation, in our framework, is designed to deliver the status of two co-training classifiers to the Q-agent. \newcite{zhang2006new} have proved that training with high-confidence examples will consequently be a process that reinforces what the current model already encodes instead of learning an accurate distribution of data space. Thus, one insight in formulating the state representation is to add some unlabeled examples with uncertainty and diversity during the training iteration. 
However, too much uncertainty will make two classifiers unstable, while too much diversity will cause the sampling bias shift towards the unlabeled dataset \cite{yeung2017learning}. In order to automatically capture this insight and select high-quality subsects during the iteration, the Q-agent needs to fully understand the distribution of the unlabeled data.

Based on the above intuition, we formulate the agent’s state using the two classifiers' probability distribution on the representative example $S_i$ of each unlabeled subset $U_i$. Suppose a $N$-class classification problem, at each time step $t$, we evaluate the probability distribution of two classifiers on $S_i$ separately. The state representation then can be defined as: 
\begin{equation}
s_t = \{P^1_1||P^2_1, P^1_2||P^2_2, ..., P^1_K||P^2_K\}_t,
\end{equation}
where $P^1_i$ and $P^2_i$ are the probability distribution of $C_1$ and $C_2$ on $S_i$ separately, and $||$ denotes the concatenation operation. $P^1_i, P^2_i \in \mathbb{R}^N$ and $P^1_i||P^2_i \in \mathbb{R}^{2N}$. Note that the state representation is re-computed at each time step $t$.

\subsubsection{Q-Network}
\begin{figure}[t]
\centering
\includegraphics[width=0.99\columnwidth]{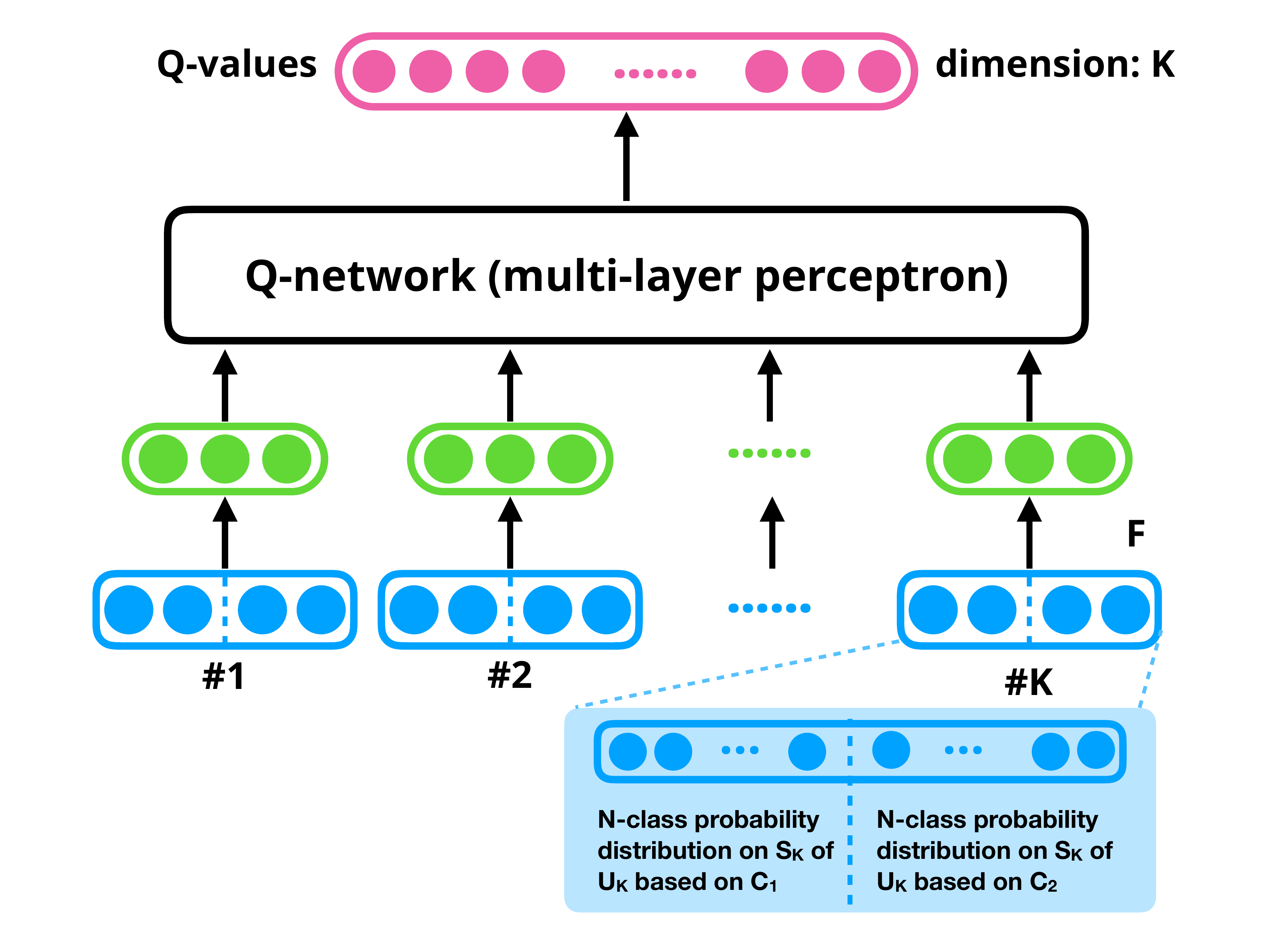}
\caption{The structure of Q-network. It chooses a unlabeled subset from $\{U_1, U_2, ...,U_K\}$ at each time step. The state representation is computed according to the two classifiers' $N$-class probability distribution on the representative example $S_i$ of each subset $U_i$.}\label{network}
\end{figure}

The agent takes an action at at time step $t$ using a policy 
\begin{equation}
a_t = \max_{a}Q(s_t, a),
\end{equation}
where $s_t$ is the state representation mentioned above. The Q-value $Q(s_t, a)$ is determined by a neural network as illustrated in Figure \ref{network}. Concretely,
\begin{equation}
z_a = \phi(\{F(P^1_1||P^2_1),..., F(P^1_K||P^2_K)\}; \theta),
\end{equation}
where the function $F$ maps state representation $P^1_i||P^2_i \in \mathbb{R}^{2N}$ into a common embedding space of $y$ dimensions, and $\phi(\cdot)$ is a multi-layer perception.

We then use
\begin{equation}
Q(s,a) = \text{softmax}(z_a)
\end{equation}
to obtain the next action.

\subsubsection{Reward Function}
The agent is trained to select the high-quality unlabeled subsets to improve the performance of the two classifier $C_1$ and $C_2$. We capture this intuition by a performance-driven reward function. At time step $t$, the reward of each classifier is defined as the change in the classifier’s accuracy after updating the unlabeled subset $U_t$:
\begin{equation}
r^1_t = \text{Acc}^1_t(L') - \text{Acc}^1_{t-1}(L'),
\end{equation}
where $\text{Acc}^1_t(L')$ is the model accuracy of $C_1$ at time step $t$ computed on the labeled validation set $L'$. Then the $r^2_t$ is defined following the similar formulation. The final reward $r_t$ is defined as:
\begin{equation*}
  r_t = 
    \begin{cases}
      r^1_t \times r^2_t & \text{if } r^1_t > 0 \text{ and } r^2_t > 0,\\
      0 & \text{otherwise}.
    \end{cases}
\end{equation*}
Note that this reward is only available during training process.

\subsection{Training and Testing}
The agent is trained with the Q-learning \cite{watkins1992q}, a standard reinforcement learning algorithm that can be used to learn policies for an agent interacting with an environment. In our Reinforced Co-Training framework, the environment is the classifier $C_1$ and $C_2$.

The Q-network parameters $\theta$ are learned by optimizing:
\begin{equation}
L_i(\theta_i) = \mathbb{E}_{s,a}[(V(\theta_{i-1})-Q(s,a;\theta_i))^2],
\end{equation}
where $i$ is an iteration of optimization and
\begin{equation}
V(\theta_{i-1}) = \mathbb{E}_{s'}[r+\gamma \max_{a'} Q(s',a';\theta_{i-1})| s,a].
\end{equation}.

We optimize it using stochastic gradient descent. The detail of the training process is shown in Algorithm \ref{training}. 

At test time, the agent and the two co-training classifiers are again run simultaneously, but without access to the labeled validation dataset. The agent selects the unlabeled subset using the learned greedy policy:
\begin{equation}
a_t = max_{a}Q(s_t, a).
\end{equation}
After obtaining two classifiers from co-training, based on the weighted voting, the final ensemble classifier $C$ is defined as:
\begin{equation}
C = \beta C_1 + (1-\beta) C_2.
\end{equation}
$\beta$ is the weighted parameter, which can be learned by maximizing the classification accuracy on the validation set.

\begin{algorithm}[t]
\caption{The algorithm of our Reinforced Co-Training method.}\label{training}
Given a set $L$ of labeled seeding training data;\\
Given a set $L'$ of labeled validation data;\\
Given $K$ subsets $\{U_1, U_2, ..., U_K\}$ of unlabeled data;\\
\For{episode $\leftarrow$ 1 \KwTo M}{
    Train $C_1$ \& $C_2$ with $L$\\
    \For {time step $t$ $\leftarrow$ 1 \KwTo T}{
        Choose the action $a_t = \max_{a}Q(s_t, a)$\\
        Use $C_1$ to label the subset $U_{a_t}$\\
        Update $C_2$ with pseudo-labeled $U_{a_t}, L$\\
        Use $C_2$ to label the subset $U_{a_t}$\\
        Update $C_1$ with pseudo-labeled $U_{a_t}, L$\\
        
        Compute the reward $r_t$ based on $L'$\\
        Compute the state representation $s_{t+1}$\\
        Update $\theta$ using $g \propto \nabla_{\theta}\mathbb{E}_{s,a}[(V(\theta_{i-1})-Q(s,a;\theta_i))^2]$
    }
}
\end{algorithm}

\section{Experiments}
\label{sec:experiments}
We evaluate our proposed Reinforced Co-training method in two settings: (1) \textbf{Clickbait detection}, where obtaining the labeled data is very time-consuming and labor-intensive in this real-world problem; (2) \textbf{Generic text classification}, where we randomly set some of the labeled data as unlabeled and train our model in a controlled setting.

\subsection{Baselines}
We compare our model with multiple baselines:
\begin{itemize*}
\item \textbf{Standard Co-Training}: Co-Training with randomly choosing unlabeled examples \cite{blum1998combining}.
\item \textbf{Performance-driven Co-Training}: The unlabeled examples are selected based on pseudo-accuracy and energy regularization \cite{zhang2006new}.
\item \textbf{CoTrade Co-Training}: The confidence of either classifier’s prediction on unlabeled examples is estimated based on specific data editing techniques, and then high-confidence examples are used to update the classifiers \cite{zhang2011cotrade}.
\item \textbf{Semi-supervised Sequence Learning (Sequence-SSL)}: The model uses an LSTM sequence autoencoder to pre-train the parameters for the later supervised learning process.\cite{dai2015semi}.
\item \textbf{Semi-supervised CNN with Region Embedding (Region-SSL)}: The model learns embeddings of small text regions from unlabeled data for integration into a supervised CNN \cite{johnson2015semi}.
\item \textbf{Adversarial Semi-supervised Learning (Adversarial-SSL)}: The model apply perturbations to word embeddings into an LSTM and pre-train the supervised models through adversarial training \cite{miyato2016adversarial}.
\end{itemize*}

\subsection{Clickbait Detection}
Clickbait is a pejorative term for web content whose headlines typically aim to make readers curious, but the documents usually have less relevance with the corresponding headlines \cite{chakraborty2016stop,potthast2017clickbait,wei2017learning}. Clickbait not only wastes the readers' time but also damages the publishers' reputation, which makes detecting clickbait become an important real-world problem.

However, most of the attempts focus on news headlines, while the relevance between headlines and context is usually ignored \cite{chen2015misleading,biyani20168,chakraborty2016stop}. Meanwhile, the labeled data is quite limited in this problem, but the unlabeled data is easily obtained from the web \cite{potthast2017clickbait}. Considering these two challenges, we utilize our Reinforced Co-training framework to tackle this problem and evaluate our method.

\subsubsection{Datasets}
We evaluate our model on a large-size clickbait dataset, Clickbait Challenge 2017 \cite{potthast2017clickbait}. The data is collected from twitter posts including tweet headlines and paragraphs, and the training and test sets are judged on a four-point scale $[0, 0.3, 0.66, 1]$ by at least five annotators. Each sample is categorized into one class based on its average scores. The clickbait detection then can be defined as a two-class classification problem, including CLICKBAIT and NON-CLICKBAIT. There also exists an unlabeled set containing large amounts of collected samples without annotation. We then split the original test set into the validation set and final test set by 50\%/50\%. The statistics of this dataset are listed in Table \ref{clickbait}.

\begin{table}[t]
\caption{\label{clickbait}Statistics of Clickbait Dataset.}
\small
\begin{tabular}{l|c|c|c}
\hline 
Dataset & \#Tweets & \#Clickbait & \#Non-Clickbait \\ 
\hline
Training & 2,495 & 762 & 1,697 \\
\hline
Validation & 9,768 & 2,380 & 7,388\\
\hline
Test & 9,770 & 2,381 & 7,389 \\
\hline
Unlabeled & 80,012 & N/A & N/A \\
\hline
\end{tabular}
\end{table}

\subsubsection{Setup}
For each document example in the clickbait dataset, naturally, we have two views, the headline and the paragraph. Thus, we construct the two classifiers in co-training based on these two views.

\textbf{Headline Classifier} The previous state-of-the-art model \cite{zhou2017clickbait} for clickbait detection uses a self-attentive bi-directional gated recurrent unit RNN (biGRU) to model the headlines of the document and train a classifier. Following the same setting, we choose self-attentive biGRU as the headline classifier in co-training.

\textbf{Paragraph Classifier} The paragraphs usually have much longer sequences than the headlines. Thus, we utilize the CNN-non-static structure in \newcite{kim:2014:EMNLP2014} as the paragraph classifier to capture the paragraph information.

Note that the other three co-training baselines also use the same classifier settings.

In our Reinforce Co-Training model, we set the number of unlabeled subsets $k$ as $80$. Considering the clickbait detection as a 2-class classification problem ($N=2$), the Q-network maps $4$-d input $P^1_i||P^2_i$ in the state representation to a $3$-d common embedding space ($y=3$), with a further hidden layer of 128 units on top. The dimension $k$ of the softmax layer is also $80$.

As for the other semi-supervised baselines, Sequence-SSL, Region-SSL and Adversarial-SSL, we concatenate the headline and the paragraph as the document and train these models directly on the document data. To better analyze the experimental results, we also implement another baseline denoted as CNN (Document), which uses the CNN structure \cite{kim:2014:EMNLP2014} to model the document with supervised learning. The CNN (Document) model is trained on the (seeding) training set and the validation set.

Following the previous researches \cite{chakraborty2016stop,potthast2017clickbait}, we use Precision, Recall and F1 Score to evaluate different models.

\subsubsection{Results}
The results of clickbait detection are shown in Table \ref{click}. From the results, we observe that: (1) Our Reinforced Co-Training model can outperform all the baselines, which indicates the capability of our methods in utilizing the unlabeled data. (2) The standard co-training is unstable due to the random data selection strategy, and the performance-driven and high-confidence data selection strategies both can improve the performance of co-training. Meanwhile, the significant improvement compared with previous co-training methods shows that the Q-agent in our model can learn a good policy to select high-quality subsets. (3) The three pre-trained based semi-supervised learning methods also show good results. We think these pre-trained based methods learn local embeddings during the unsupervised training, which may help them to recognize some important patterns in clickbait detection. (4) The self-attentive biGRU trained only on headlines of the labeled set actually show surprisingly good performance on clickbait detection, which demonstrates that most clickbait documents have obvious patterns in the headline field. The reason why CNN (Document) fails to capture these patterns may be that the concatenation of headlines and paragraphs dilutes these features. But for those cases without obvious patterns in the headline, our results demonstrate that the paragraph information is still a good supplement to detection.

\begin{table}[t]
\small
\begin{center}
\begin{tabular}{l|c|c|c}
\hline 
Methods & \textbf{Prec.} & \textbf{Recall} & \textbf{F1 Score} \\
\hline
Self-attentive biGRU & 0.683 & 0.649 & 0.665\\
CNN (Document) & 0.537 & 0.474 & 0.503 \\
\hline
Standard Co-Training & 0.418 & 0.433 & 0.425 \\
Performance Co-Training & 0.581 & 0.629 & 0.604 \\
CoTrade Co-Training & 0.609 & 0.637 & 0.623 \\
\hline
Sequence-SSL & 0.595 & 0.589 & 0.592 \\
Region-SSL & 0.674 & 0.652 & 0.663 \\
Adversarial-SSL & 0.698 & \bf{0.691} & 0.694 \\
\hline
Reinforced Co-Training & \bf{0.709} & 0.684 & \bf{0.696}\\
\hline
\end{tabular}
\end{center}
\caption{\label{click}The experimental results on clickbait dataset. Prec.: precision.}
\end{table}

\subsubsection{Algorithm Robustness}
\label{subsubsec:robust}
Previous studies \cite{morimoto2001robust,henderson2017deep} show that reinforcement learning-based methods usually lack robustness and are sensitive to the seeding sets and pre-trained steps. Thus, we design an experiment to detect whether our learned data section policy is sensitive to the (seeding) training set. First, based on our original data partition, we train our reinforcement learning framework to learn a Q-agent. During the test time, instead of using the same seeding set when doing comparative experiments, we randomly sample other $10$ seeding sets from the labeled dataset and learn $10$ classifiers based without re-training the Q-agent (data selection policy). Note that the validation set is not available during the co-training period of the test time. Finally, we evaluate these $10$ classifiers using the same metric. The results are shown in Table \ref{click_robust}.

\begin{table}[t]
\small
\begin{center}
\begin{tabular}{c|c|c|c|c}
\hline 
 & \textbf{Best} & \textbf{Worst} & \textbf{Average} & \textbf{STDDEV} \\
\hline
F1 Score & 0.708 & 0.685 & 0.692 & 0.0068 \\
\hline
\end{tabular}
\end{center}
\caption{\label{click_robust}The robustness analysis on clickbait dataset.}
\end{table}

The results demonstrate that our learning algorithm is robust to different (seeding) training sets, which indicates that the Q-agent in our model can learn a good and robust data selection policy to select high-quality unlabeled subsets to help the co-training process.

\subsection{Generic Text Classification}
Generic text classification is a classic problem for natural language processing, where one needs to categorized documents into pre-defined classes \cite{kim:2014:EMNLP2014,zhang2015character,johnson2015semi,johnson2016supervised,xiao2016efficient,miyato2016adversarial}. We evaluate our model on generic text classification problem to study our method in a controlled setting.

\subsubsection{Datasets}
Following the settings in \newcite{zhang2015character}, we use large-scale datasets to train and test our model. To maintain the two-view setting of the co-training method, we choose the following two datasets. The original annotated training set is then split into three sets, 10\% labeled training set, 10\% labeled validation set and 80\% unlabeled set. The original proportion of different classes remains the same after the partition. The statistics of these two datasets are listed in Table \ref{text}.

{\bf AG's news corpus}. The AG’s corpus of news articles is obtained from the web and each sample has the title and description fields.

{\bf DBpedia ontology dataset}. This dataset is constructed by picking 14 non-overlapping classes from DBpedia 2014. Each sample contains the title and abstract of a Wikipedia article.

\begin{table}[tb]
\caption{\label{text}Statistics of the Text Classification Datasets.}
\small
\begin{tabular}{l|c|c}
\hline
Dataset & AG's News & DBpedia \\
\hline
\#Classes & 4 & 14 \\
\hline
\#Training & 12,000 & 56,000 \\
\hline
\#Validation & 12,000 & 56,000 \\
\hline
\#Test & 7,600 & 70,000 \\
\hline
\#Unlabeled & 96,000 & 448,000 \\
\hline
\end{tabular}
\end{table}

\subsubsection{Setup}
For each document example in the above two datasets, naturally we have two views, the headline and the paragraph. Similar to clickbait detection, we also construct the two classifiers in co-training based on these two views. Following the \cite{kim:2014:EMNLP2014}, we set both the headline classifier and the paragraph classifier as the CNN-non-static model. Owing to that fact that the original datasets are fully labeled, we implement two other baselines: (1) CNN (Training+Validation), which is supervised trained on the partitioned training and validation sets; (2) CNN (All) which is supervised trained on the original ($100\%$) dataset.

For AG's News dataset, we set the number of unlabeled subsets $k$ as $96$. The number of classes $N=4$, and thus the Q-network maps $8$-d input $P^1_i||P^2_i$ in the state representation to a $5$-d common embedding space ($y=5$), with a further hidden layer of 128 units on top. The dimension $k$ of the softmax layer is also $96$. As for DBpedia dataset, $k=224, N=14$, and $y=10$,.

Following the previous researches \cite{kim:2014:EMNLP2014}, we use test error rate (\%) to evaluate different models.

\subsubsection{Results}
The results of generic text classification are shown in Table \ref{text_results}. From the results, we can observe that: (1) Our Reinforced Co-Training model outperforms all the real semi-supervised baselines on two generic text classification datasets, which indicates that our method is consistent in different tasks. (2) The CNN (All) and Adversarial-SSL trained on all the original labeled data perform best, which indicates there is still an obvious gap between semi-supervised methods and full-supervised methods.

\begin{table}[tb]
\caption{\label{text_results}The experimental results on generic text classification datasets. * Adversarial-SSL is trained on full labeled data after pre-training.}
\small
\begin{tabular}{l|c|c}
\hline 
Methods & \textbf{AG's News} & \textbf{DBpedia} \\
\hline
CNN (Training+Validation) & 28.32\% & 9.53\%\\
CNN (All) & 8.69\%  & 0.91\%\\
\hline
Standard Co-Training & 26.52\% & 7.66\%\\
Performance Co-Training & 21.73\% & 5.84\%\\
CoTrade Co-Training & 19.06\% & 5.12\%\\
\hline
Sequence-SSL & 19.54\% & 4.64\% \\
Region-SSL & 18.27\% & 3.76\%\\
Adversarial-SSL & $8.45\%^*$ & $0.89\%^*$\\
\hline
Reinforced Co-Training & \textbf{16.64\%} & \textbf{2.45\%}\\
\hline
\end{tabular}
\end{table}

\subsubsection{Algorithm Robustness}
\label{subsubsec:robust_2}
Similar to Section \ref{subsubsec:robust}, we evaluate whether our learned data section policy is sensitive to the different partitions and (seeding) training sets. First, based on our original data partition ($10\%/10\%/80\%$), we train our reinforcement learning framework. During the test time, we randomly sample other $10$ data partitions instead of the one used in comparative experiments, and learn $10$ ensemble classifiers based on the learned Q-agent. Note that after sample different data partitions, we will also reprocess the unlabeled sets as described in Section \ref{subsec:partition}. We then evaluate these $10$ classifiers using the same metric. The results are shown in Table \ref{text_robust}.

\begin{table}[t]
\caption{\label{text_robust}The robustness analysis on generic text classification. Metric: test error rate (\%).}
\small
\begin{tabular}{l|c|c|c|c}
\hline 
Datasets & \textbf{Best} & \textbf{Worst} & \textbf{Average} & \textbf{STDDEV} \\
\hline
AG's News & 14.78 & 17.96 & 16.62 & 1.36 \\
\hline
DBPedia & 2.18 & 4.06 & 2.75 & 0.94 \\
\hline
\end{tabular}
\end{table}

The results demonstrate that our learning algorithm is robust to different (seeding) training sets and partitions of the unlabeled set, which again indicates that the Q-agent in our model is able to learn a good and robust data selection policy to select high-quality unlabeled subsets to help the co-training process.

\subsection{Discussion about Stability}
Previous studies \cite{zhang2014comprehensive,reimers-gurevych:2017:EMNLP2017} show that neural networks can be unstable even with the same training parameters on the same training data. As for our cases, when the two classifiers are initialized with different labeled seeding sets, they can be very unstable. However, after enough iterations with the properly selected unlabeled data, the performance would be stable generally.

Usually, the more substantial labeled training datasets will lead to more stable models. However, the problem is that the AG’s News and DBpedia have 4 and 14 classes separately, while the Clickbait dataset only has 2 classes. That means the numbers of each class in AG’s News, DBPedia and Clickbait actually are the same order of magnitude. Meanwhile, in our co-training setting, the prediction error is easy to accumulate because the two classifiers bootstrap the performance of each other. The classification could be harder with the increase of classes. Based on these reasons, the stability does not show a very strong correlation with the size of datasets in our experiments of Section~\ref{subsubsec:robust} and~\ref{subsubsec:robust_2}.

\section{Conclusion and Future Work}
\label{sec:conclusion}
In this paper, we propose a novel method, Reinforced Co-Training, for training classifiers by utilizing both the labeled and unlabeled data. The Q-agent in our model can learn a good data selection policy to select high-quality unlabeled data for co-training. We evaluate our models on two tasks, clickbait detection and generic text classification. Experimental results show that our model can outperform other semi-supervised baselines, especially those conventional co-training methods. We also test the Q-agent and prove that the learned data selection policy is robust to different seeding sets and data partitions. 

For future studies, we will investigate the data selection policies of other semi-supervised methods and try to learn these policies automatically. We also plan to extend our method to multi-source classification cases and utilize the multi-agent communication environment to boost the classification performance.

\section*{Acknowledgments}
\label{sec:ack}
The authors would like to thank the anonymous reviewers for their thoughtful comments. The work was supported by an unrestricted gift from Bytedance (Toutiao).

\bibliographystyle{acl_natbib}
\bibliography{naaclhlt2018}

\end{document}